\documentclass[a4paper,fleqn]{cas-dc}

\usepackage[authoryear]{natbib}
\usepackage{cleveref}
\usepackage{subfigure}
\usepackage{algorithm}  
\usepackage{algorithmicx}  
\usepackage{algpseudocode} 
\def\tsc#1{\csdef{#1}{\textsc{\lowercase{#1}}\xspace}}
\tsc{WGM}
\tsc{QE}
\tsc{EP}
\tsc{PMS}
\tsc{BEC}
\tsc{DE}

\begin{document}
\let\WriteBookmarks\relax
\let\printorcid\relax
\def\floatpagepagefraction{1}
\def\textpagefraction{.001}
\shorttitle{}
\shortauthors{Yan Xiao et~al.}
\title [mode = title]{Demand Forecasting in Bike-sharing Systems Based on A Multiple
Spatiotemporal Fusion Network}                      



\author[1]{\color{black}Xiao Yan}
\author[1]{\color{black}Gang Kou}
\author[1]{\color{black}Feng Xiao}
\author[1]{\color{black}Dapeng Zhang }
\author[1]{\color{black}Xianghua Gan\corref{cor1}}
\ead{ganx@swufe.edu.cn}

\address[1]{
  School of Business Administration,
  Faculty of Business Administration,
  Southwestern University of Finance and Economics, Chengdu, China.
}

\cortext[cor1]{Corresponding author}

\begin{abstract}
  Bike-sharing systems (BSSs) have become increasingly popular
	around the globe and have
	attracted a wide range of research interests.
	In this paper, the demand forecasting problem in BSSs is studied.
	Spatial and temporal features are critical for
	demand forecasting in BSSs,
	but it is challenging to extract spatiotemporal dynamics.
	Another challenge is to capture the relations
	between spatiotemporal dynamics
	and external factors, such as weather, day-of-week, and time-of-day.
	To address these challenges,
	we propose a multiple spatiotemporal fusion network named 
	MSTF-Net.
	MSTF-Net consists of multiple spatiotemporal blocks: 3D convolutional network (3D-CNN) blocks, eidetic 3D convolutional long short-term memory networks (E3D-LSTM) blocks, and fully-connected (FC) blocks.
	Specifically, 3D-CNN blocks highlight extracting short-term spatiotemporal dependence in each fragment (i.e., closeness, period, and trend); E3D-LSTM blocks further extract long-term
	spatiotemporal dependence over all fragments; FC blocks extract nonlinear correlations of external factors.
	Finally, the latent representations of E3D-LSTM and FC blocks are fused to obtain the final prediction.
	For two real-world datasets,
	it is shown that MSTF-Net outperforms seven state-of-the-art models.
\end{abstract}



\maketitle

\section{Introduction} \label{introduction}

The earliest bike-sharing system (BSS), i.e., the White Bike Sharing Manifesto \citep{floret2014bike}, dates to the 1960s.
Examples of similar systems include
Citybike in New York City,
Divvy in Chicago,
Capital Bikeshare in Washington D.C.,
Ecobici in the Mexico City,
and Velib in France.
These systems are called station-based
since a user has to rent or return a bike to some fixed bike station
and may have to end the trip early or waste time returning the bike.
Owing to GPS technology,
station-free bike-sharing systems (SFBSSs) have become prevalent.
They alleviate the inconvenience of renting and returning bikes
at station-based BSSs.
With a GPS-enabled smartphone,
a user can locate a bike nearby and park it at her/his convenience.
Recently,
SFBSSs have become increasingly popular
in many cities around the World,
and have introduced a way of living lighter,
consuming less,
and protecting the environment.
As an example,
station-free BSSs dominate the market in China,
and millions of bikes from such systems have flooded most large cities.
The dataset used in this paper is from a global bike-sharing company named Mobike (acquired by a company named Meitun in 2018),
which provides riding services to more than 200 million users in more than 200 cities in 19 countries around the world as of 2018.

A crucial problem for BSSs is the imbalance of bikes
owing to one-way trips between regions and customer-demand uncertainties.
This problem has attracted researchers and practitioners
\citep{rudloff2014modeling,dell2014bike, alvarez2016optimizing, schuijbroek2017inventory, a1, Li2018a}.
An adequate solution to the problem
relies on the forecasting of customer demand.
It is this issue of demand forecasting that is the focus of the present paper.

A few papers \citep[e.g.,][]{b1,b2} have been
devoted to forecasting problems in BSSs,
which are reviewed in more detail in \Cref{literature}.
Recently, deep-learning approaches have been used
in forecasting problems in large-scale bike-sharing networks.
A common method is to generate traffic flow videos for each region of a city from transactions, each of
which consists of the start time, end time, start location, and end location. The videos are then fed into deep-learning models.
\citet{c1} employ a 3D convolutional neural network (3D-CNN)
to effectively capture the spatiotemporal dependence jointly
from low- to high-level layers for traffic flow data.
Since the 3D-CNN architecture captures long-term relations by sampling and assembling,
it does not perform well in discovering the spatiotemporal dependence between cause and effect.
\citet{c2} uses a convolutional long short-term memory network (ConvLSTM)
to extract the spatial dependence by the encapsulated 2D convolutions and the long-term relations by LSTM, and shows better performance in traffic flow prediction.
However, the ConvLSTM network only establishes temporal connections on the high-level features at the top layer,
while leaving the spatial and temporal correlations on the low-level features not fully exploited.

Motivated by this research,
an objective of the present study is
to combine the advantages of both 3D-CNN and ConvLSTM
in spatiotemporal predictive learning.
A newly developed deep-learning model \citep{d3} is used,
which is named the eidetic 3D convolutional long short-term memory network (E3D-LSTM),
to process the short-term frame dependence and long-term high-level relations.
The encapsulated 3D-CNN makes local perceptrons of LSTM motion-aware and
enables the memory cell to store better short-term features. In addition, the present memory state can interact with its historical records via a gate-controlled self-attention module for long-term spatiotemporal relations.
Demand for bikes in a BSS can be directly affected by external factors such as weather, day-of-week, and time-of-day.
For example, fewer people tend to ride bikes on a rainy day.
As another example,
bike demands around central business districts are usually high on weekdays,
but low on weekends.
Fusing the external factors and the spatiotemporal data is the second objective of this study.

Letting inflow (outflow) of the region be the number of bikes that enter (leave) the region during a time interval,
inflow and outflow videos are first generated from transaction data by partitioning a city into a grid map based on the longitude and latitude, where
a grid denotes a region.
We propose a multiple
spatiotemporal fusion network named MSTF-Net.
The videos are processed by 3D-CNN Encoder layers to extract short-term dependence and obtain high-dimensional feature maps.
Then, the feature maps are fed into E3D-LSTM layers to extract  long-term spatiotemporal interaction as hidden states.
Next, the hidden states are decoded by 3D-CNN Decoder layers to obtain a latent representation, and a $1 \times 1$ 2D-CNN layer is implemented to map the latent representation to inflow-outflow channel.
Meanwhile, fully-connected layers are implemented to extract non-linear correlations for external factors.
Finally, we add the outputs of the 2D-CNN layer and the fully-connected layers to obtain the final prediction.

The contributions of this paper are summarized as the following.
\begin{itemize}
	\item We propose a multiple spatiotemporal fusion network named MSTF-Net that
	can enhance the prediction capacity in bike-sharing demand forecasting problem.
	\item In MSTF-Net, 3D-CNN blocks highlight extracting short-term spatiotemporal dependence in each fragment (i.e., closeness, period, and trend); E3D-LSTM blocks extract long-term spatiotemporal dependence over all fragments; FC blocks extract nonlinear correlations of external factors.
	\item 
	We conduct ablation studies for the multiple blocks design. It is shown that MSTF-Net significantly
	performs better than the pure 3D-CNN or E3D-LSTM models.
	\item The spatial and temporal correlations are visualized
    and the significances of spatial and temporal dependence
    in demand forecasting are validated.
\end{itemize}


\section{Literature review} \label{literature}
A transaction record consists of spatial features and temporal features.
Forecasting models, such as the Linear Regressor, SVR, and Adaboost Regressor,
take these records in vector form as model inputs directly.
Papers that use these models are briefly reviewed first.
Instead of using these records directly,
some researchers first generate flow videos from transaction data,
and then use deep-learning models such as 2D-CNN, 3D-CNN, and ConvLSTM models.
These papers are closer to our research and are the focus of this review.
\citet{hong2011traffic} presents a  model that combines the seasonal support vector regression model with chaotic simulated annealing algorithm (SSVRCSA), to forecast inter-urban traffic flow. 
\citet{giot2014predicting} use the Adaboost Regressor, Ridge Regression, SVR, Random Forest Regressor, and Gradient Boosting Regressor to forecast demand in bike-sharing systems; it is found that most regressors are sensitive to overfitting by multiple experiments.
\citet{ashqar2017modeling} first proposed a bipartite clustering algorithm to cluster bike stations into groups, and then use a Gradient Boosting Regression Tree (GBRT) to predict the inflow and outflow of each station in the groups.
\citet{xu2018station} investigated the mobility pattern of SFBSSs, and employed long short-term memory  (LSTMs) neural networks to forecast the demand.
\citet{Negahban2019} proposed a methodology combining simulation, bootstrapping, and subset selection that uses the useful partial information in every bike
pickup/drop-off observation to estimate the true demand in bike-sharing
systems.

The following papers describe the transformation of transaction data to flow videos
and use deep-learning models for forecasting.
\citet{Zhang2016} proposed a deep-learning model based on a 2D convolutional neural network (2D-CNN) to simultaneously extract spatial
 dependence, temporal closeness, periods, and trends in bike-sharing systems.
\citet{li2018origin} employed 2D-CNN and LSTM models to capture spatial and temporal dependence, respectively, for forecasting distributions of origin and destination.
\citet{c1} employed a 3D convolutional neural network (3D-CNN), which redesigns the inner mechanism by considering the temporal dimension based on a 2D-CNN, to forecast outflow and inflow in bike-sharing systems, and showed that a 3D-CNN model has better performance than a 2D-CNN model.
A 3D-CNN model can extract features from both the spatial and temporal dimensions by performing 3D convolutions, thereby capturing the motion information encoded in multiple adjacent frames.
 \citet{c2} employed a convolutional long short-term memory network (ConvLSTM), a deep combination of 2D-CNN and LSTM, to forecast bike distribution, and showed that the ConvLSTM model has better performance than 2D-CNN and LSTM models.
 ConvLSTM extends LSTM to have 2D convolutional structures in both the input-to-state and state-to-state transitions, which can process the spatial and temporal dependence in one approach.

Demand prediction problems in bike-sharing systems were studied in all the aforementioned papers.
In online ride-hailing systems, the problem of demand prediction is also important,
and has attracted numerous researchers.
Herein, only two closely related papers are reviewed.
\citet{Ke2017} proposed a new fusion deep-learning architecture based on ConvLSTM to fuse exogenous variables such as weather, day-of-week, and time-of-day to predict the demand of each region.
\citet{Zhang2019} proposed a fully convolutional neural network architecture based on 3D-CNN, and employed locally connected 2D convolutional layers to predict the demand of each region.
Recently, in active traffic management systems,  \citet{zhang2020network} proposed a hybrid forecasting approach by integrating 3D-CNN with ensemble empirical mode decomposition to forecast traffic speed.

\section{Preliminaries} \label{preliminaries}
\begin{figure}[h] \centering
    \subfigure[Outflow matrix by one hour] {
	\label{fig:a}
    \includegraphics[width=0.4\columnwidth]{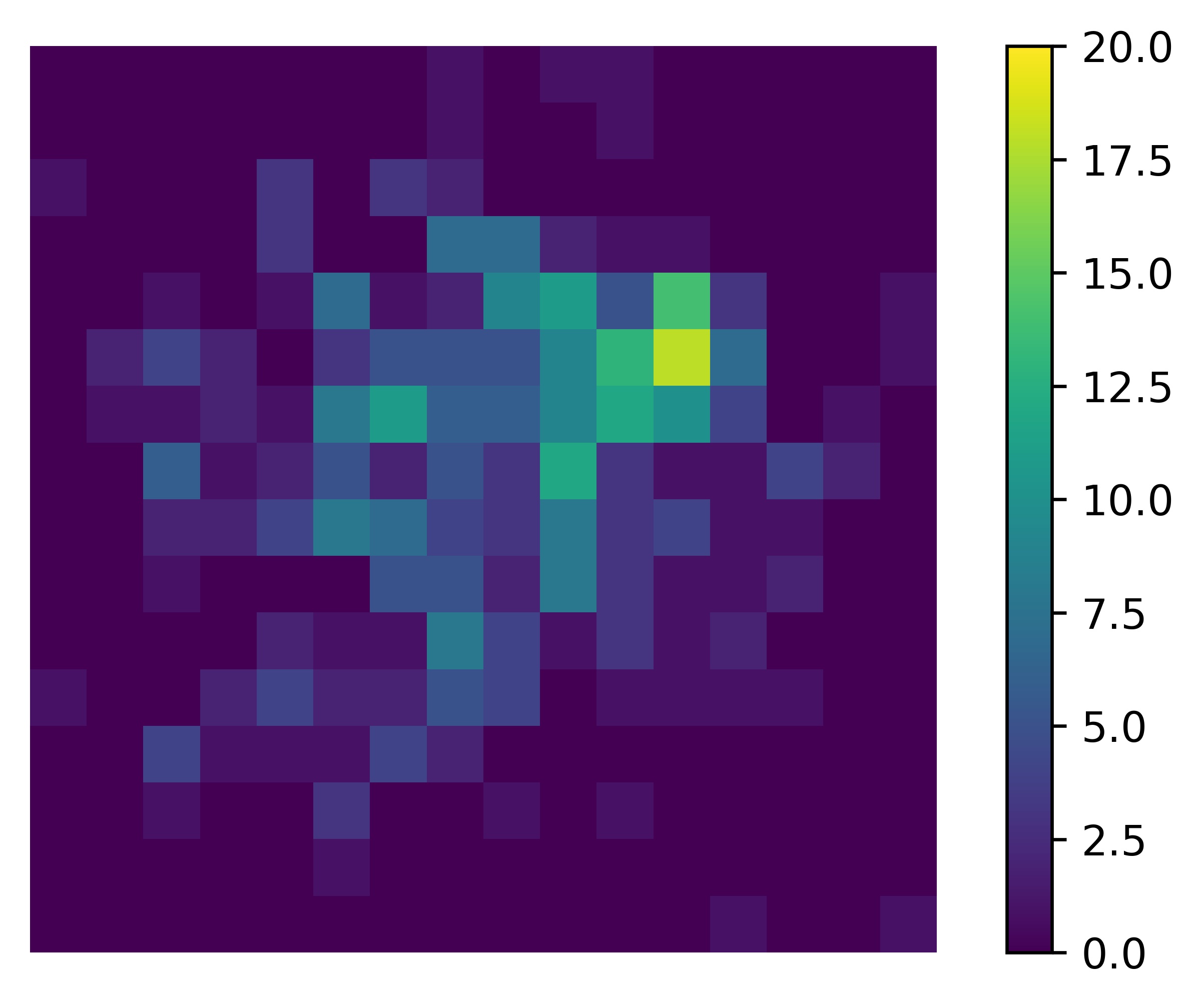}
    }
    \subfigure[Inflow and outflow] {
    \includegraphics[width=0.35\columnwidth]{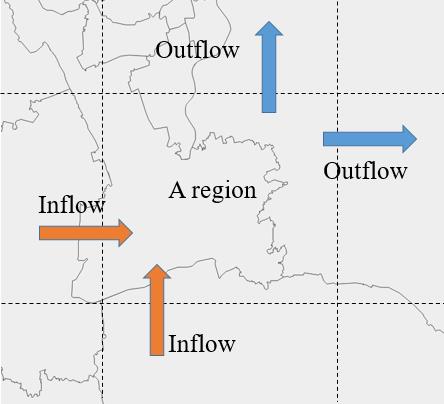}
    }
    \caption{Regions in Shanghai}
    \label{Regions in Shanghai}
    \end{figure}

We first define the problem of demand forecasting in bike-sharing systems.
\newtheorem{mydef}{Definition}
\newtheorem{mypro}{Problem}
\begin{mydef}
\textbf{(Region and time partition \citep{Zhang2016})}
The city area is partitioned into $I \times J$ grids uniformly according to coordinates, as shown in \Cref{Regions in Shanghai}.
\end{mydef}


\begin{mydef}
	\textbf{(Inflow/outflow \citep{Zhang2016})}
		Let $R$ be a collection of trajectories at the $t^{th}$ time interval. For a grid $(i, j)$ that lies at the $i^{th}$ row and the $j^{th}$ column, the inflows and outflows at time interval $t$ are defined respectively as:
		\end{mydef}
	
	\begin{align*}
		d_{t}^{in, i, j} &=\sum_{Tr \in R} \mid\left\{k \geq 1 \mid  g_{k}^{end} \in(i, j)\right\}  \\
		d_{t}^{out, i, j} &=\sum_{Tr \in R} \mid\left\{k \geq 1 \mid
		g_{k}^{start} \in(i, j) \right\},
		\end{align*}

	where $\operatorname{Tr}: g_{1} \rightarrow g_{2} \rightarrow \cdots
	\rightarrow g_{|T r|}$ is a trajectory in $R$, and $g_{k}^{start}$,
	$g_{k}^{end}$ are the geospatial coordinate; $g_{k}^{start} \in(i, j)$,
	$g_{k}^{end} \in(i, j)$ mean the trajectory start or end in the grid $(i, j)$,
	note that the trajectory can start and end in the same region,
	$| \cdot  |$ denotes the cardinality of a set.
	
	At time interval $t$, inflows and outflows in all $I \times  J$ regions can be
	denoted by a tensor $D_{t} \in \mathbb{R}^{2 \times I \times J}$ where $\left(D_{t}\right)_{0, i, j}=d_{t}^{i n, i, j},\left(D_{t}\right)_{1, i, j}=d_{t}^{\text {out }, i, j}$.
	The outflow matrix is shown in \Cref{fig:a}.







\begin{mypro}
	Predict $D_{n}$ given historical observations $\{ D_{t}| t=0, \cdots , n-1 \} $ and external factors such as weather conditions, wind speed, temperature, and day-of-week.
	\end{mypro}

\section{Method} \label{method}

\begin{figure*}[htbp]
	\centering
	\includegraphics[width=0.9\textwidth]{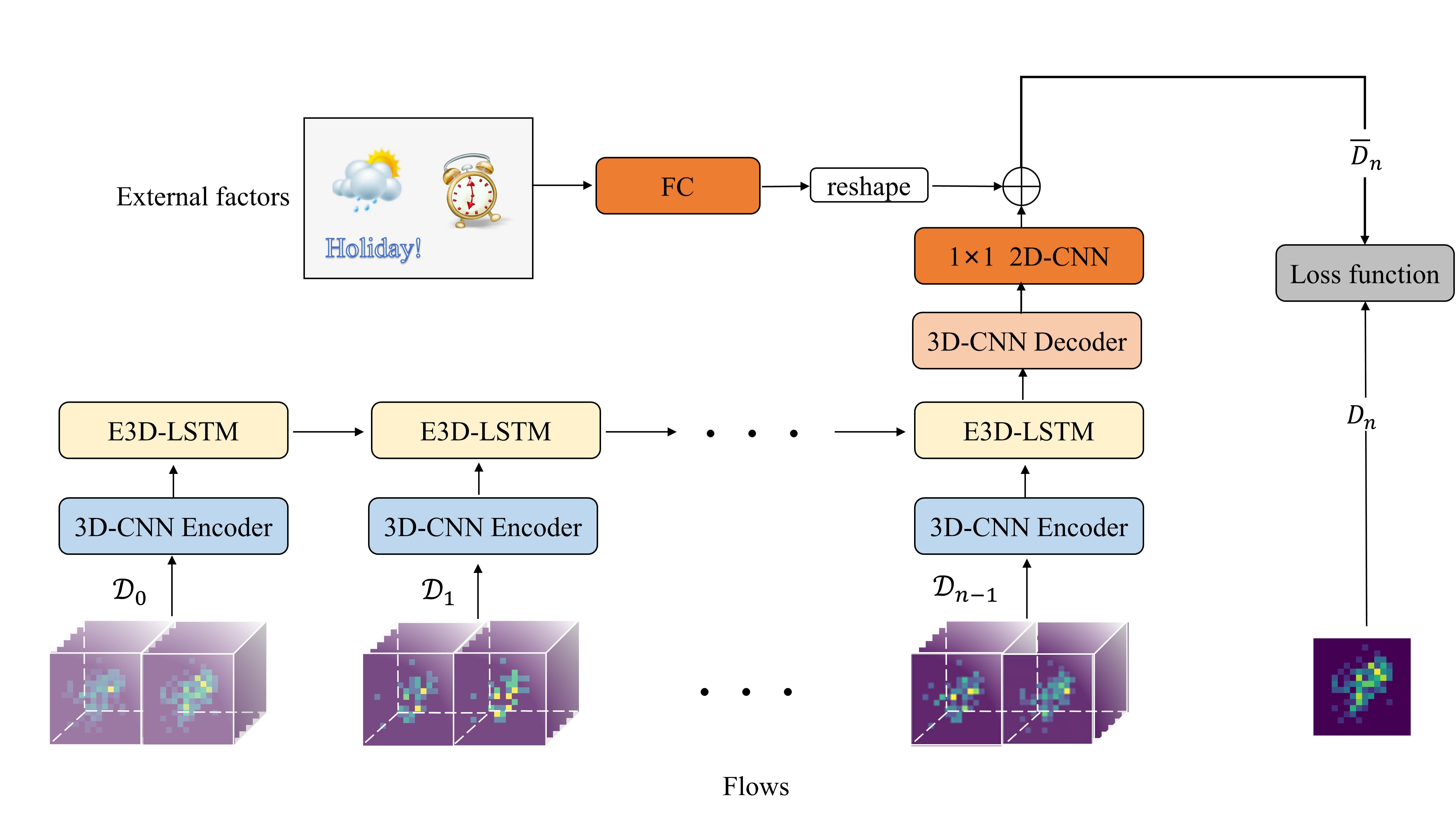}
	\caption{Architecture of MSTF-Net}
	\label{fig 5}
	\end{figure*}

This section presents our new model named
MSTF-Net.
We introduce the architecture of MSTF-Net,
which is illustrated in \Cref{fig 5}.
First, 3D-CNN Encoder layers are implemented to process historical observations 
to extract short-term dependence and obtain high-dimension feature maps.
Then the feature maps are directly fed into E3D-LSTM layers to further extract the long-term spatiotemporal interaction as hidden states.
Next, the E3D-LSTM hidden states are decoded by 3D-CNN Decoder layers and one $1 \times  1$ 2D-CNN layer
to get a latent representation $\mathcal{Z}$.
Meanwhile,
FC blocks are implemented
to extract non-linear correlations $\mathcal{V}$ for external factors.
Finally,
we fuse $\mathcal{Z}$ and $\mathcal{V}$  to obtain our final prediction $D_{n}$.


\subsection{Input} \label{input_videos}
We sample historical flow videos from recent time to near history and distant history according to three corresponding temporal views: closeness, period, and trend.
We select hours, daily, and weekly as the key timesteps to construct the three views.
For each of temporal views, we fetch a list of key timesteps’ flow matrices and concatenated them, to construct the input as:

	\begin{align*}
		&D_{closeness}=\left[D_{t-1}, D_{t-2}, \cdots, D_{t-l_{r}}\right] \in \mathbb{R}^{N \times C \times l_{r}} \\
		&D_{period}=\left[D_{t-p_{d}}, D_{t-2 p_{d}}, \cdots, D_{t-l_{d} * p_{d}}\right] \in \mathbb{R}^{N \times C \times l_{d}} \\
		&D_{trend}=\left[D_{t-p_{w}}, D_{t-2 p_{w}}, \cdots, D_{t-l_{w} * p_{w}}\right] \in \mathbb{R}^{N \times C \times l_{w}} \\
		&\{D_{t}| t=0, \cdots , n-1 \}=\left[D_{closeness}, D_{period}, D_{trend} \right] ,
		\end{align*}

where  $l_{r},l_{d},l_{w}$ are input lengths of  hours, daily, and weekly, $p_{d}$, $p_{w}$ are daily and weekly periods.

\subsection{Structures for spatiotemporal variables} \label{spa}

\begin{figure}[ht]
	\centering
	\includegraphics[width=0.35\textwidth]{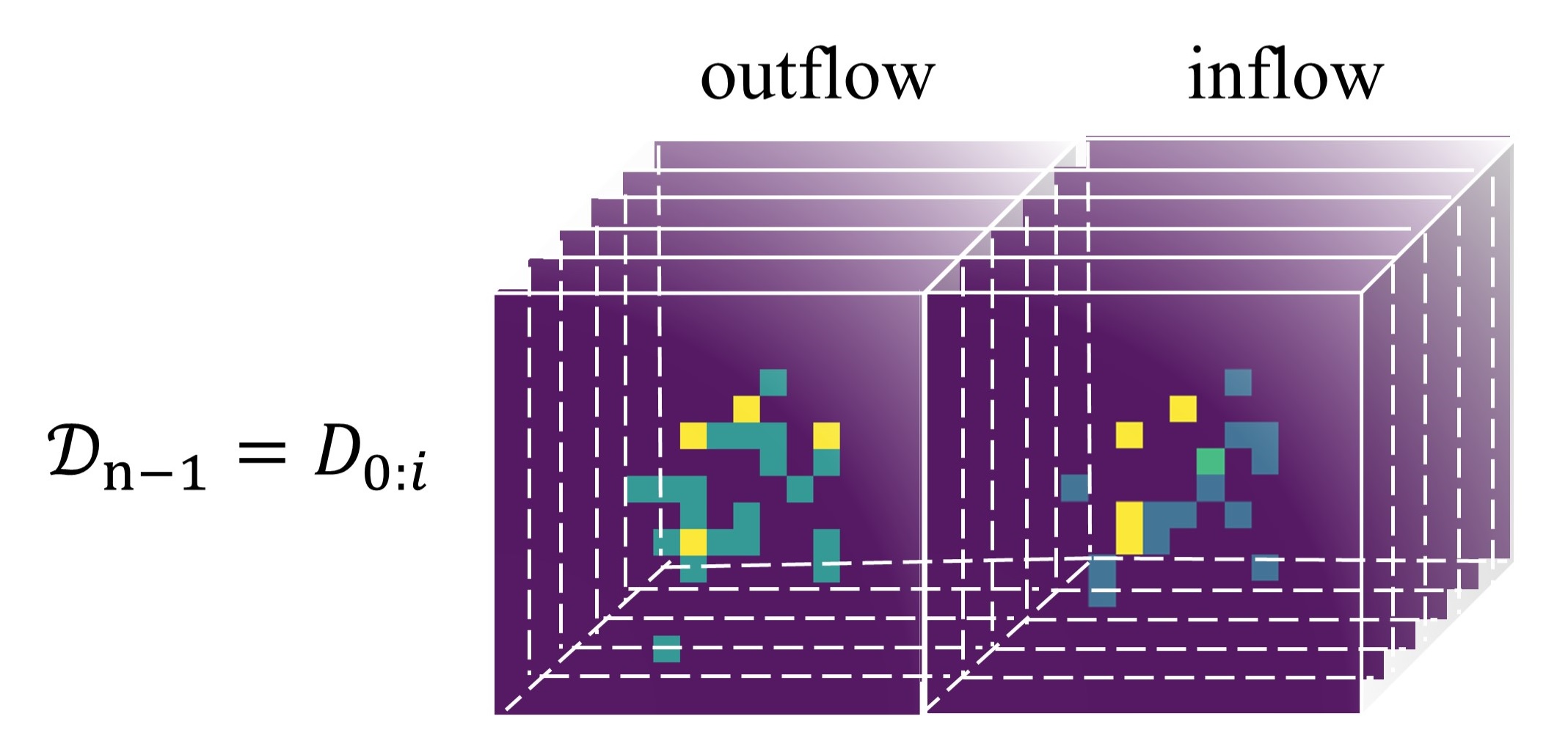}
	\caption{A fragment of flow videos on MoBike dataset. $i$ is the number of frames in a fragment.}
	\label{fig video}
	\end{figure}

For flow videos, each fragment is encoded by a shared 3D-CNN Encoder
to extract short-term dependence obtain high-dimensional feature maps.
The motivation for the implementation of 3D-CNN in encoding representation
is that 3D-CNN is suitable to extract short-term appearance and local motions in a consecutive short-term period:

\begin{align*}
	&(\mathcal{Y}_{0}, \mathcal{Y}_{1}, \ldots, \mathcal{Y}_{n-1}) =
	RELU ( \mathcal{F}_{l}^{Encoder} \ldots 
	\\ & \mathcal{F}_{1}^{Encoder}
	(\mathcal{D}_{0}, \mathcal{D}_{1}, \ldots, \mathcal{D}_{n-1})),
\end{align*}

where $\mathcal{D}_{n-1}=D_{0:i}$ is the $n-1$th fragment that consists of $i$ frames, as show in \Cref{fig video}, $RELU$ is activation function, $l$ is the number of layers, $n$ is the number of fragments, 
and $\mathcal{F}^{Encoder}$ is the 3D-CNN Encoder.

Next, the outputs
$(\mathcal{Y}_{0}, \mathcal{Y}_{1}, \ldots, \mathcal{Y}_{n-1})$
are fed into the E3D-LSTM blocks that integrate 3D convolutions into LSTM to capture long short-term dependence.
The purpose for implementation of the E3D-LSTM is to learn better representations for both short-term frame dependence and long-term high-level relations.
Specifically, the encapsulated 3D convolution makes local perceptrons of units
motion-aware and enables the memory cell to store better short-term features.
For long-term relations, the present memory state interacts with its historical records via a gate-controlled self-attention \\ LSTM module:

	\begin{align*}
	&(\mathcal{U}_{0}, \mathcal{U}_{1}, \ldots, \mathcal{U}_{n-1})=RELU(\mathcal{F}_{l^{'}}^{E3D-LSTM} \ldots \\& \mathcal{F}_{1}^{E3D-LSTM}(\mathcal{Y}_{0}, \mathcal{Y}_{1}, \ldots, \mathcal{Y}_{n-1})),
	\end{align*}

where $\mathcal{F}^{E3D-LSTM}$ is the E3D-LSTM block.

\begin{figure*}[htbp]
	\centering
	\includegraphics[width=0.9\textwidth]{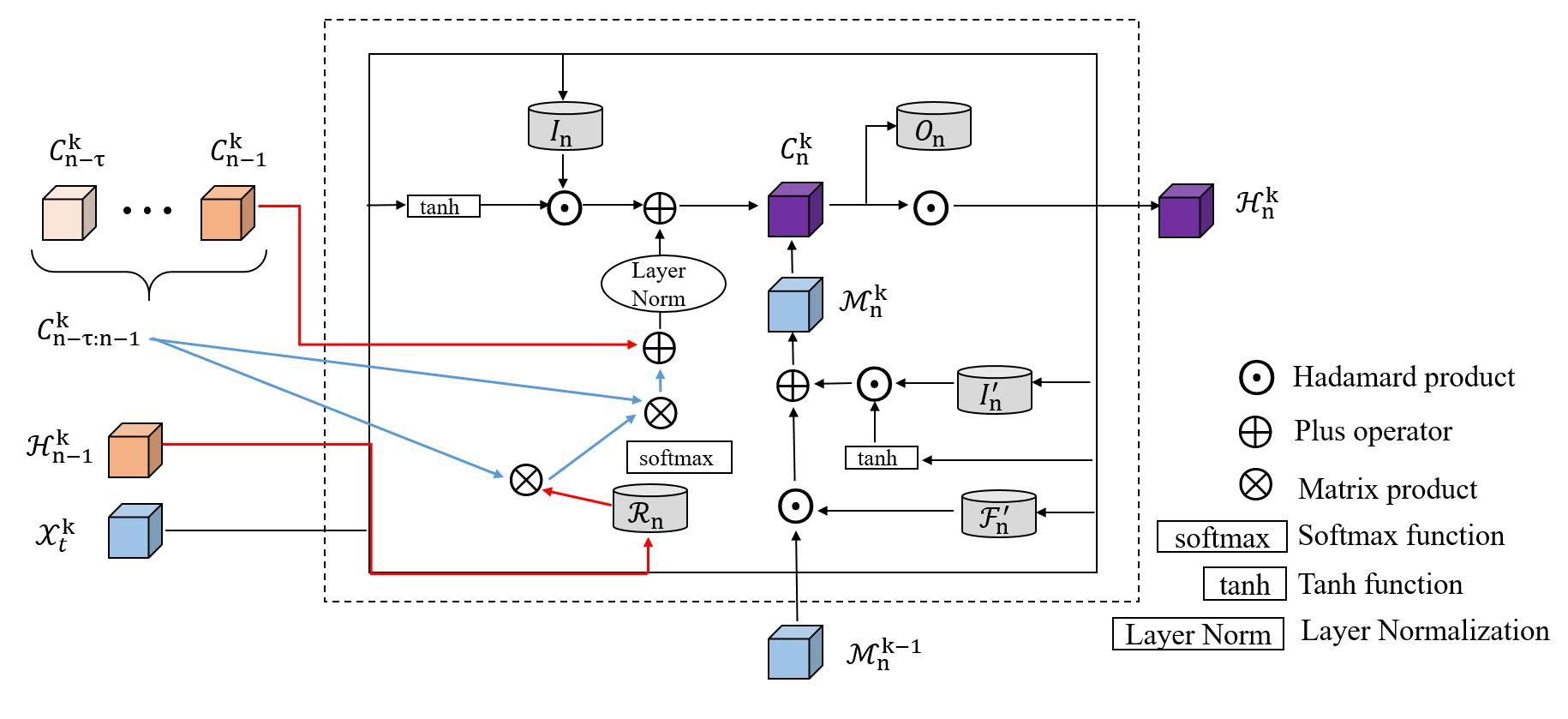}
	\caption{The inner structure of a E3D-LSTM layer}
	\label{fig e3d-lstm}
	\end{figure*}

The inner architecture of the E3D-LSTM is illustrated in \Cref{fig e3d-lstm}, where the red arrows indicate short-term information flow and the blue arrows denote long-term information flow in hour dimension.
The architecture can be divided into the temporal part and the spatiotemporal part.

In the temporal part, recurrent 3D convolution as motion-aware perceptrons to extract 10-minute appearance and local motions in continuous space-time fields and store them in a small spatiotemporal volume.
 A memory RECALL mechanism is designed for
 the recurrent transition function of the memory states to capture hourly interactions:

	\begin{align*}
		&\mathcal{R}_{n}=\sigma\left(W_{x r} * \mathcal{X}_{n}+W_{h r} * \mathcal{H}_{n-1}^{k}+b_{r}\right) \\
		&\mathcal{I}_{n}=\sigma\left(W_{x i} * \mathcal{X}_{n}+W_{h i} * \mathcal{H}_{n-1}^{k}+b_{i}\right) \\
		&\mathcal{G}_{n}=\tanh \left(W_{x g} * \mathcal{X}_{n}+W_{h g} * \mathcal{H}_{n-1}^{k}+b_{g}\right) \\
		&\operatorname{RECALL}\left(\mathcal{R}_{n}, \mathcal{C}_{n-\tau: n-1}^{k}\right)=\operatorname{softmax}(\mathcal{R}_{n} \\ &\cdot\left(\mathcal{C}_{n-\tau: n-1}^{k}\right)^{\top}) \cdot \mathcal{C}_{n-\tau: n-1}^{k} \\
		&\mathcal{C}_{n}^{k}=\mathcal{I}_{n} \odot \mathcal{G}_{n}+\text { LayerNorm }(\mathcal{C}_{n-1}^{k}+ \\ & \operatorname{RECALL}(\mathcal{R}_{n}, \mathcal{C}_{n-\tau: n-1}^{k})),
	\end{align*}

where $n$ is the hour, $\sigma$ is the sigmoid function, $\ast$ is 3D convolution operation, $\odot$ is Hadamard product, $\text { LayerNorm }$ is layer normalization, $\cdot$ is matrix product after reshaping the recall gate $\mathcal{R}_{n}$ and long-term memory states $\mathcal{C}_{n-\tau: n-1}^{k}$ into matrixes,
$b_r,b_i,b_g$ are intercept parameters.
$\mathcal{G}_{n}$ is an interaction of the current input $\mathcal{X}_{n}$ and the previous short-term memory state $\mathcal{H}_{n-1}^{k}$,
 and $\mathcal{I}_{n}$ is an input gate controls which parts of $\mathcal{G}_{n}$ should be added to the long-term state like standard LSTM.
$\mathcal{R}_{n}$ is a recall gate, acting as memory access instructions, controls where and what to attend in historical memory records.
The RECALL function is implemented as an attentive
 module to compute the relationship between the encoded local patterns and the whole long-term memory space to get the current long-term memory state $\mathcal{C}_{n}^{k}$.
 The hyper-parameter $\tau$ means how many historical memory states are attended by the recall gate $\mathcal{R}_{n}$.
 On the other hand, the RECALL function is a self-attention mechanism that is used to evoke past memories from distant timestamps for memorizing and distilling useful information from what has been perceived (i.e., $\tau=n-1$ in our study).

 With the updated memory state $\mathcal{C}_{n}^{k}$, the hidden states are:

	\begin{align*}
	&\mathcal{I}_{n}^{\prime}=\sigma\left(W_{x i}^{\prime} * \mathcal{X}_{n}+W_{m i} * \mathcal{M}_{n}^{k-1}+b_{i}^{\prime}\right) \\
	&\mathcal{G}_{n}^{\prime}=\tanh \left(W_{x g}^{\prime} * \mathcal{X}_{n}+W_{m g} * \mathcal{M}_{n}^{k-1}+b_{g}^{\prime}\right) \\
	&\mathcal{F}_{n}^{\prime}=\sigma\left(W_{x f}^{\prime} * \mathcal{X}_{n}+W_{m f} * \mathcal{M}_{n}^{k-1}+b_{f}^{\prime}\right) \\
	&\mathcal{M}_{n}^{k}=\mathcal{I}_{n}^{\prime} \odot \mathcal{G}_{n}^{\prime}+\mathcal{F}_{n}^{\prime} \odot \mathcal{M}_{n}^{k-1} \\
	&\mathcal{O}_{n}=\sigma(W_{x o} * \mathcal{X}_{n}+W_{h o} * \mathcal{H}_{n-1}^{k}+W_{c o} * \\ &\mathcal{C}_{n}^{k}+W_{m o} * \mathcal{M}_{n}^{k}+b_{o}) \\
	&\mathcal{H}_{n}^{k}=\mathcal{O}_{n} \odot \tanh \left(W_{1 \times 1 \times 1} *\left[\mathcal{C}_{n}^{k}, \mathcal{M}_{n}^{k}\right]\right),
	\end{align*}

where $W_{1 \times 1 \times 1}$ is the $1 \times 1 \times 1$ convolutions for the transformation of the channel number.
In the spatiotemporal part (see Appendix for details),
$\mathcal{I}_{n}^{\prime},\mathcal{G}_{n}^{\prime}$ are the gate structures similar to $\mathcal{I}_{n},\mathcal{G}_{n}$ mentioned before,
the forget gate $\mathcal{F}_{n}^{\prime}$ controls which parts of the
long-term state should be erased, $\mathcal{M}_{n}^{k}$ is the previous spatiotemporal memory states.
Finally, the output gate $\mathcal{O}_{n}$ controls which parts of the long-term state should be read and output short-term memory state $\mathcal{H}_{n}^{k}$ at the current timestamp.
For simplicity, the number of layers $k$ is omitted in the current layer.
Note that there are two E3D-LSTM layers in MSTF-Net, so $\mathcal{X}_{n}^k= \mathcal{Y}_{n}$ and $\mathcal{H}_{n}^k= \mathcal{U}_{n}$ when $k=2$:
\begin{equation*}
	(\mathcal{U}_{0},\mathcal{U}_{1} \ldots \mathcal{U}_{n-1} )= (\mathcal{H}_{0}^k, \mathcal{H}_{1}^k \ldots \mathcal{H}_{n-1}^k ).
\end{equation*}

Then, 3D-CNN Decoder layers are employed to get the latent representation $\mathcal{Z}$:
\begin{equation*}
	\mathcal{Z}=\mathcal{F}_{l^{''}}^{Decoder} \ldots \mathcal{F}_{1}^{Decoder}( \mathcal{U}_{0}, \mathcal{U}_{1}, \ldots, \mathcal{U}_{n-1}),
\end{equation*}
where $\mathcal{F}^{Decoder}$ is the 3D-CNN layer.
The Decoder extracts higher-level feature maps.

\subsection{ Structures for temporal variables}\label{stv}

Temporal variables include current external factors $Ext_{n-1}$ such as
weather condition, day-of-week, temperature, wind speed, and time-of-day.
We use fully-connected layers to extract the non-linear correlations $\mathcal{V}$ between them.
This is represented by
\begin{equation*}
	\mathcal{V}=\mathcal{F}_{l^{'''}}^{FC} \ldots \mathcal{F}_{1}^{FC}(Ext_{n-1}),
\end{equation*}
where $\mathcal{F}^{FC}$ is the fully-connected layers.

\subsection{Fusion}\label{fusion}
We process $\mathcal{Z}$ and $\mathcal{V}$ to obtain
final prediction $\overline{D}_{n}$:
\begin{equation*}
	\overline{D}_{n}=F^{2D-CNN}(\mathcal{Z})+Reshape^{vector\rightarrow tensor}(\mathcal{V})
\end{equation*}
where
$F^{2D-CNN}$ is a $1 \times 1$ 2D-CNN layer that maps $\mathcal{Z}$ back to inflow-outflow channel.
$Reshape$ is a function that 
transforms vectors back to videos, denoted by $Reshape^{vector\rightarrow tensor}: R^{M}\rightarrow R^{T \times C \times I \times J} $, $M=T*C*I*J$.

\subsection{Training algorithm}\label{ssub:training_algorithm}

During the training process of MSTF-Net,
the object is to minimize the Mean Squared Error (MSE) between the real flow $D_{n}$ and the estimated flow $\overline{D}_{n+1}$.
The objective function is formulated by
\begin{equation}\label{loss}
	\min _{w, b}\|D_{n}-\overline{D}_{n}\|_{2}^{2}.
\end{equation}

The training steps are illustrated in Algorithm \ref{alg:Framwork}.
\begin{algorithm}
	\caption{MSTF-Net training}
	\label{alg:Framwork}
	\begin{algorithmic}[1]
	  	\Require
			The historical observations of flow videos, \{$D_{0} \ldots D_{m}$ \};

			The external factors, $Ext_{m}$;

			The look-back windows, $d$;

	  	\Ensure
		The learned parameters of MSTF-Net;
	  	\State initialize a null set: $L$
		\For{time-series $d \leqq t \leqq m$}
			\State $\mathcal{I}_{t-1} \leftarrow \{D_{t-d}\ldots D_{t-1}\} $
		\State A training dataset $\{\{\mathcal{I}_{t-1}, Ext_{t-1}\},D_{t}\}$ is put into $L$
		\EndFor
		\State Initialize all the weighted and intercept parameters
		\Repeat
		\State Randomly sample a batch $L_{s}$ from $L$
		\State Minimizing the objective function shown in \Cref{loss} to obtain the parameters within $L_{s}$
		\Until{convergence criterion met}
	\end{algorithmic}
  \end{algorithm}

\section{Experiment} \label{experiment}
In \Cref{dataset}, we present the datasets.
In \Cref{step}, we set up experiments.
In \Cref{compare}, we show the results and analysis.
In \Cref{parameter sensitivity}, we show the parameter sensitivity.
in \Cref{ablation study}, we show the ablation study.
In \Cref{visua}, we visualize the results.

\subsection{Datasets}\label{dataset}
 We use two datasets, including the trajectory data of station-free sharing bike in Shanghai and station-based sharing bike in New York City (NYC).

\textbf{MoBike:} The trajectory data is station-free sharing bike GPS data of MoBike for Shanghai from 1st Aug. 2018 to 31th Aug. 2018 about 100 thousand trajectories.  We  partition Shanghai into $16 \times 16$ regions.  
For  MoBike, we just sample historical flow videos from recent time and select 10-minutes as the key timestep, because the time span  is just one month, which are not enough to be sampled from near history and distant history. 

\textbf{NYCBike:} The trajectory data is station-based sharing bike GPS data  for New York City (NYC) from 1st Jan. 2018 to 31th Dec. 2020, about 56 million trajectories.
We partition NYC into $12 \times 16$ regions.

For MoBike, we choose data from the last four days as the test set, all data before that as the training set. The last four days of training set is chosen as validation set. 
For NYCBike, we choose data from the last four weeks as the test set, all data before that as the training set. The last four weeks of training set is chosen as validation set. 

\textbf{Exploring the spatial and temporal correlations.}
Take MoBike  as an example, we explore correlations between the flow (outflow or inflow)
in a given region at the $n$th hour and the spatiotemporal variables
ahead of the $n$th hour by employing the Pearson correlation, given by
\begin{equation*}
	\operatorname{Corr}(Y, X)=\frac{E[(Y-E(Y))^{\prime}(X-E(X))]}{E[(Y-E(Y))^{2}] E[(X-E(X))^{2}]},
\end{equation*}
where $Y$ and $Z$ are two random variables with the same number of observations.

\begin{figure*}[htbp]
	\centering
	\begin{minipage}[t]{0.48\textwidth}
	\centering
	\includegraphics[width=1\textwidth]{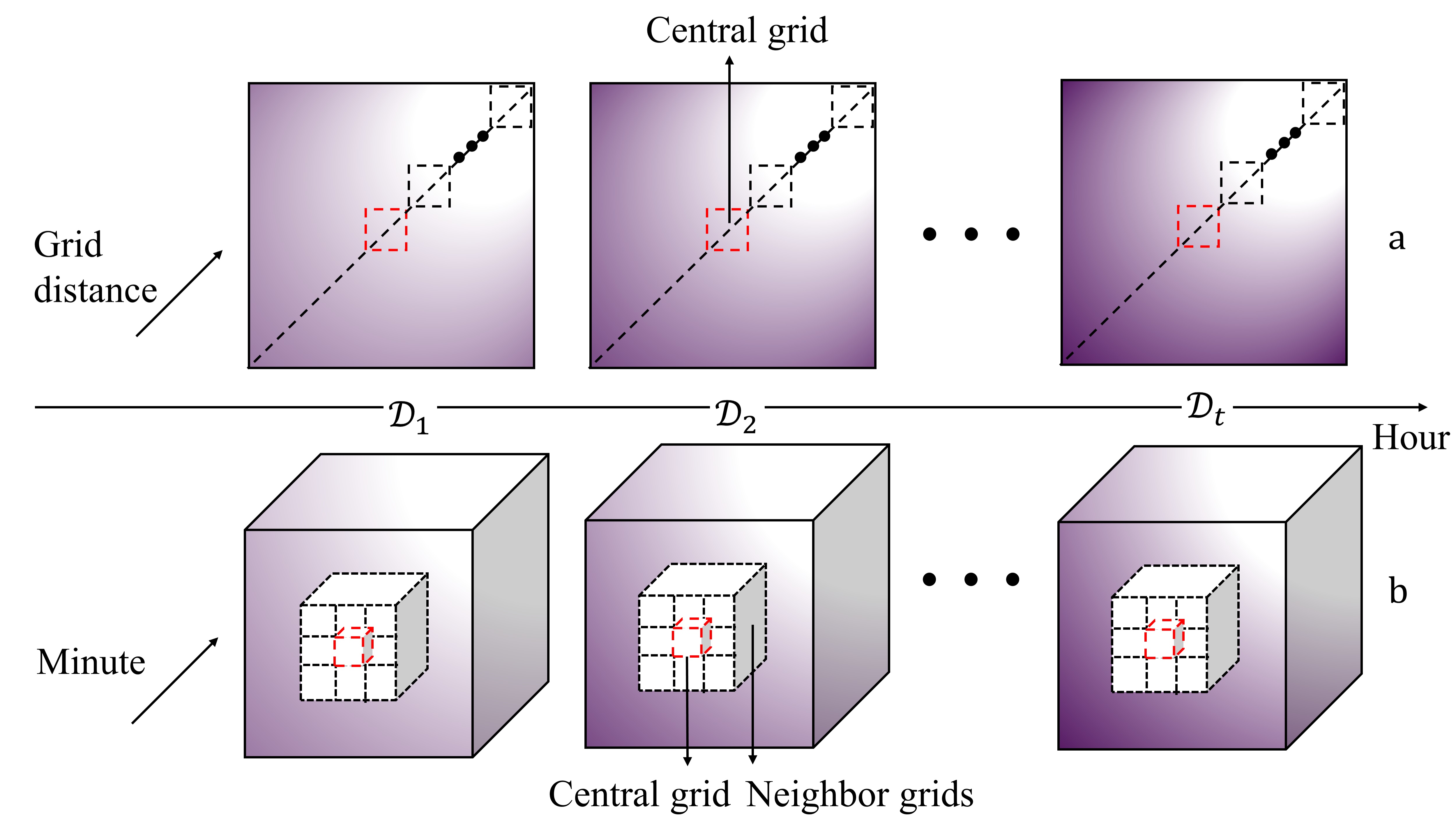}
	\caption{Central grid and Neighbor grids}
	\label{fig vector}
	\end{minipage}
	\begin{minipage}[t]{0.48\textwidth}
	\centering
	\includegraphics[width=1\textwidth]{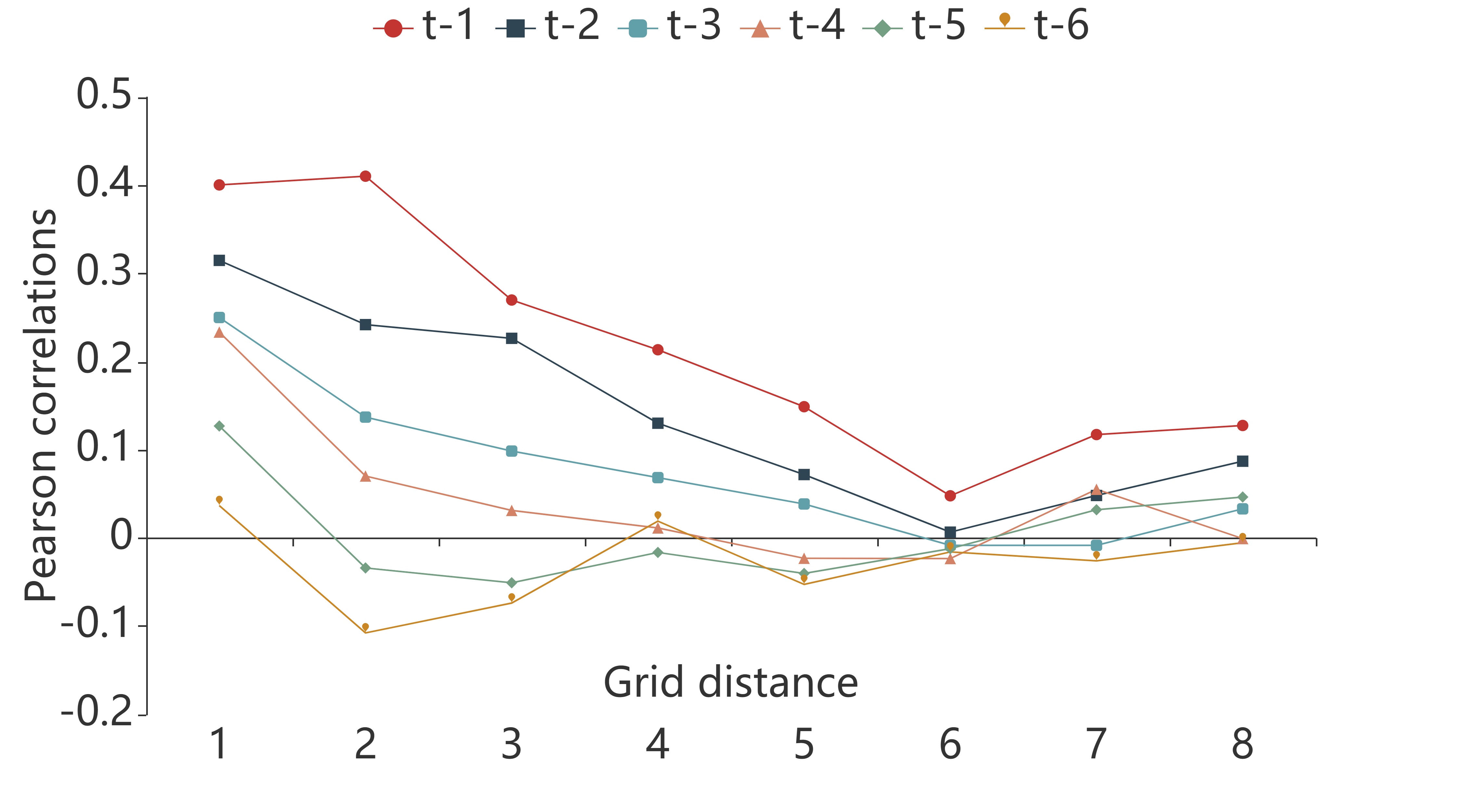}
	\caption{Pearson correlations across time and space}
	\label{fig corre}
	\end{minipage}
	\end{figure*}

Take outflow in central grid $(7,7)$ as an example,
as shown in \Cref{fig vector}.
The grid distance of grid $(i,j)$ and $(7,7)$ equals $i-7$ or $j-7$.
\Cref{fig corre} shows correlations between  grid $(7,7)$ and  grid $(i,j)$ from 6 hours ago.
Overall, correlations drop gradually with the increase of  grid distance, which indicates that there exit strong spatial correlations between grid $(7,7)$ and its neighbor regions.
On the other hand, it is not surprising that variables with shorter look-back windows have higher correlations.
This correlation analysis of MoBike provides evidence that  spatial and temporal dependence exist among  spatiotemporal variables.

\begin{figure*}[htbp]
\centering
\begin{minipage}[t]{0.4\textwidth}
\centering
\includegraphics[width=1\textwidth]{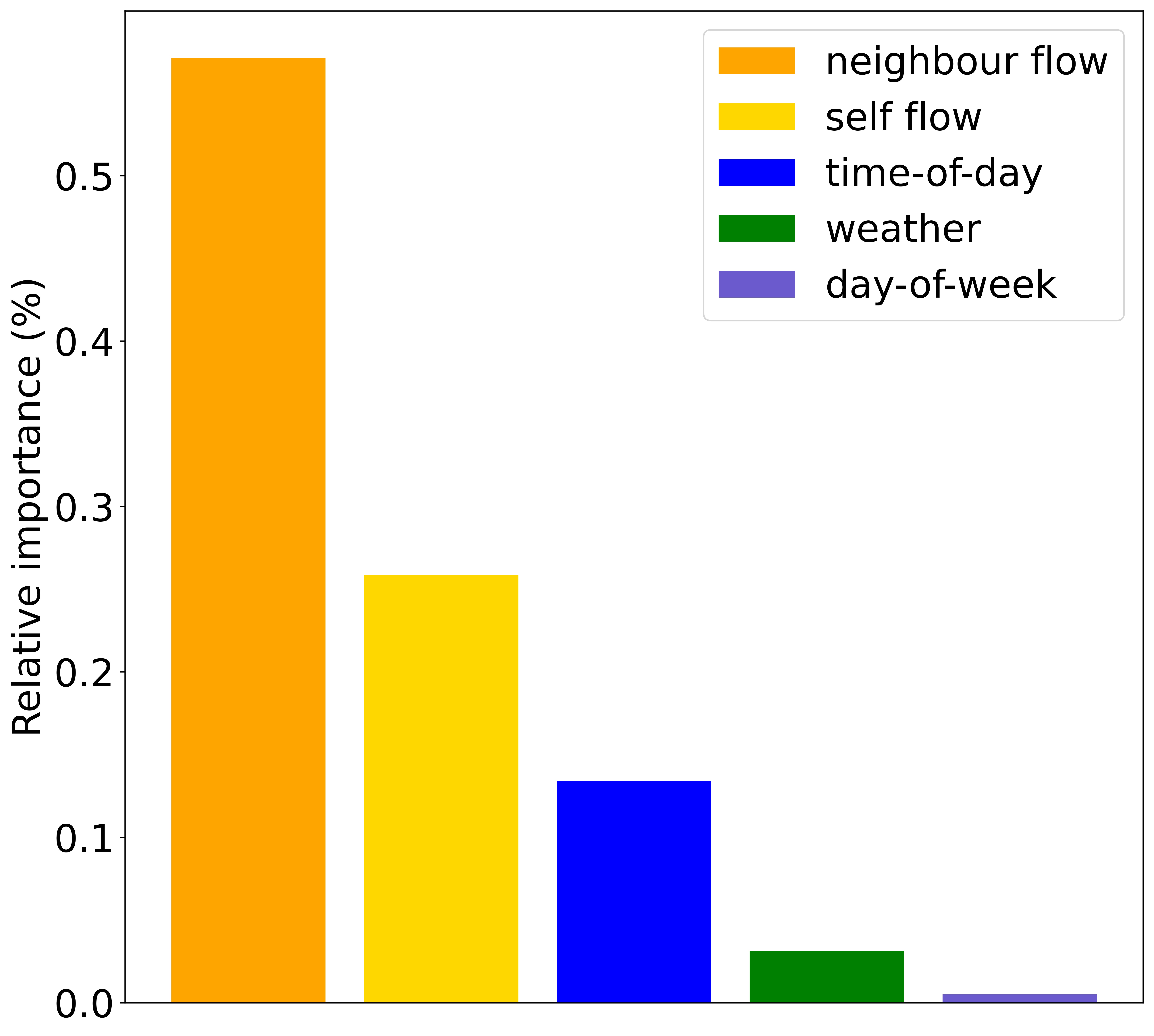}
\caption{Variable importance ranking}
\label{fig importance_variable}
\end{minipage}
\begin{minipage}[t]{0.48\textwidth}
\centering
\includegraphics[width=1\textwidth]{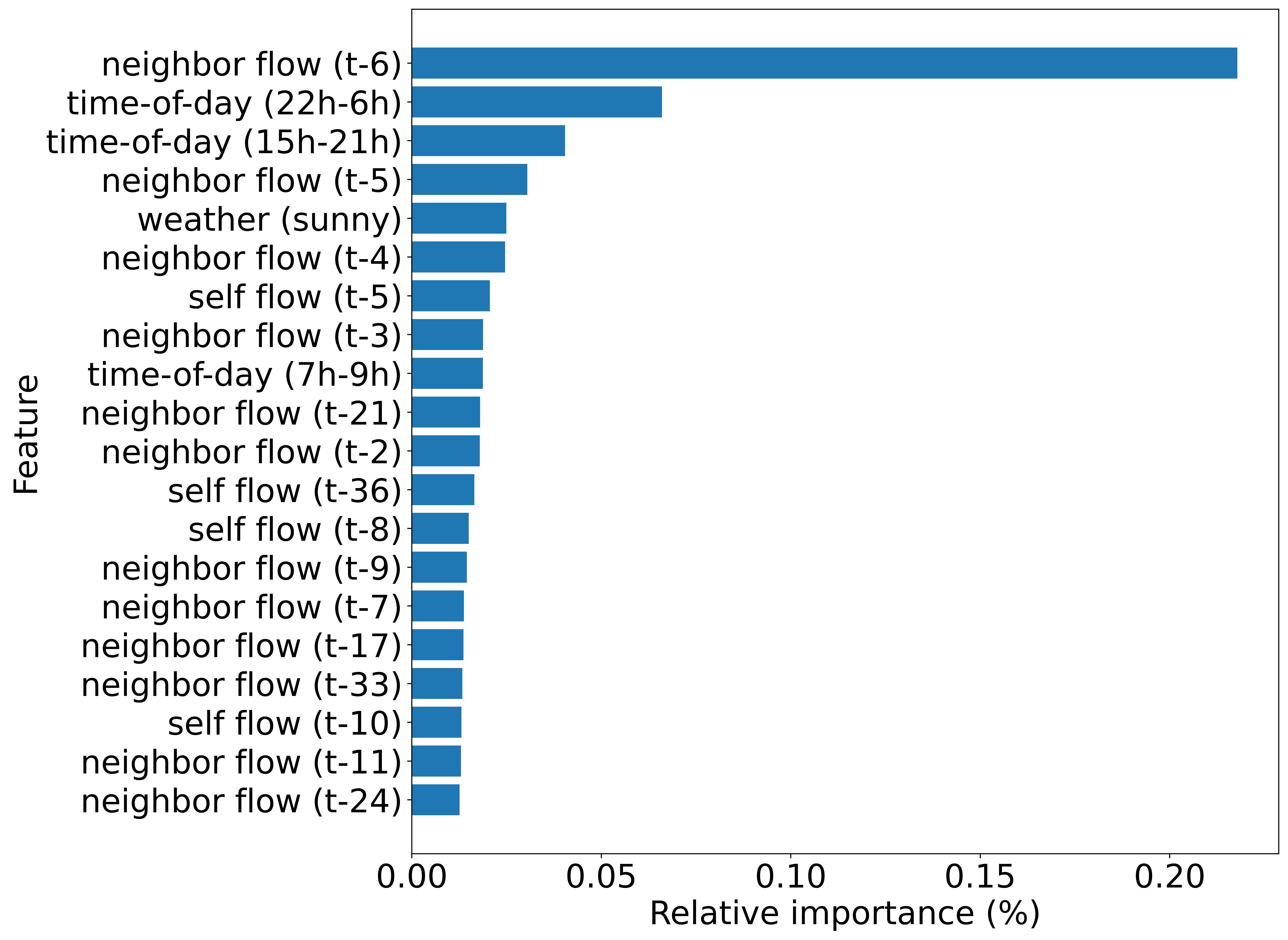}
\caption{ Feature importance (weight)}
\label{fig importance}
\end{minipage}
\end{figure*}

\textbf{Exploring the feature importance.}
Take MoBike as an example, to measure the feature importance of the spatiotemporal and temporal variables,
we model the flow (outflow or inflow) in a given region and its neighbor regions with XGBoost \citep{chen2016xgboost}, which is a gradient boosting tree model.

Take outflow in central grid $(7,7)$ as an example, as shown in \Cref{fig vector} (b),
we show the top 20 critical features measured by weight (i.e., the number of times a feature appears) and information gain (i.e., the average information gain a feature appears),
which are generally used to evaluate the features in general.
As shown in \Cref{fig importance_variable} and \Cref{fig importance}, the neighbor flow $(t-n)$ means the average outflow of the neighbor grids at the $t-n$ time interval and
the flow $(t-n)$ means the self-outflow in the grid (7,7) at the $t-n$ time interval, $n \in \{1,2,...36\}$ and time interval is 10 minutes.

\Cref{fig importance_variable} shows the variable importance partitioned by category. It can be observed that the neighbor flow, self flow, and time-of-day are the dominating factors. Other variables, such as weather and day-of-week have less contribution (less than 5\%) to the prediction.
\Cref{fig importance} shows the top 20 important features. It can be found that
the importance of neighbor flow $(t-6)$ significantly surpasses the other features for prediction, measured by the information gain.
The top features of neighbor flow and self flow are located in shorter time intervals.
The other variables such as weather, time-of-day, and day-of-week are not as  important as expected. One reason is that variables are not independent, and  there exists multicollinearity between the other variables and  the flow.
The other variables have influenced the flow before measuring the importance. For example, the small historical  flows on a rainy day generally also predict a little flow even though we don't know today is  a rainy day.




\subsection{Experimental setups and Baselines} \label{step}
\begin{figure*}[htbp]
	\centering
	\includegraphics[width=0.9\textwidth]{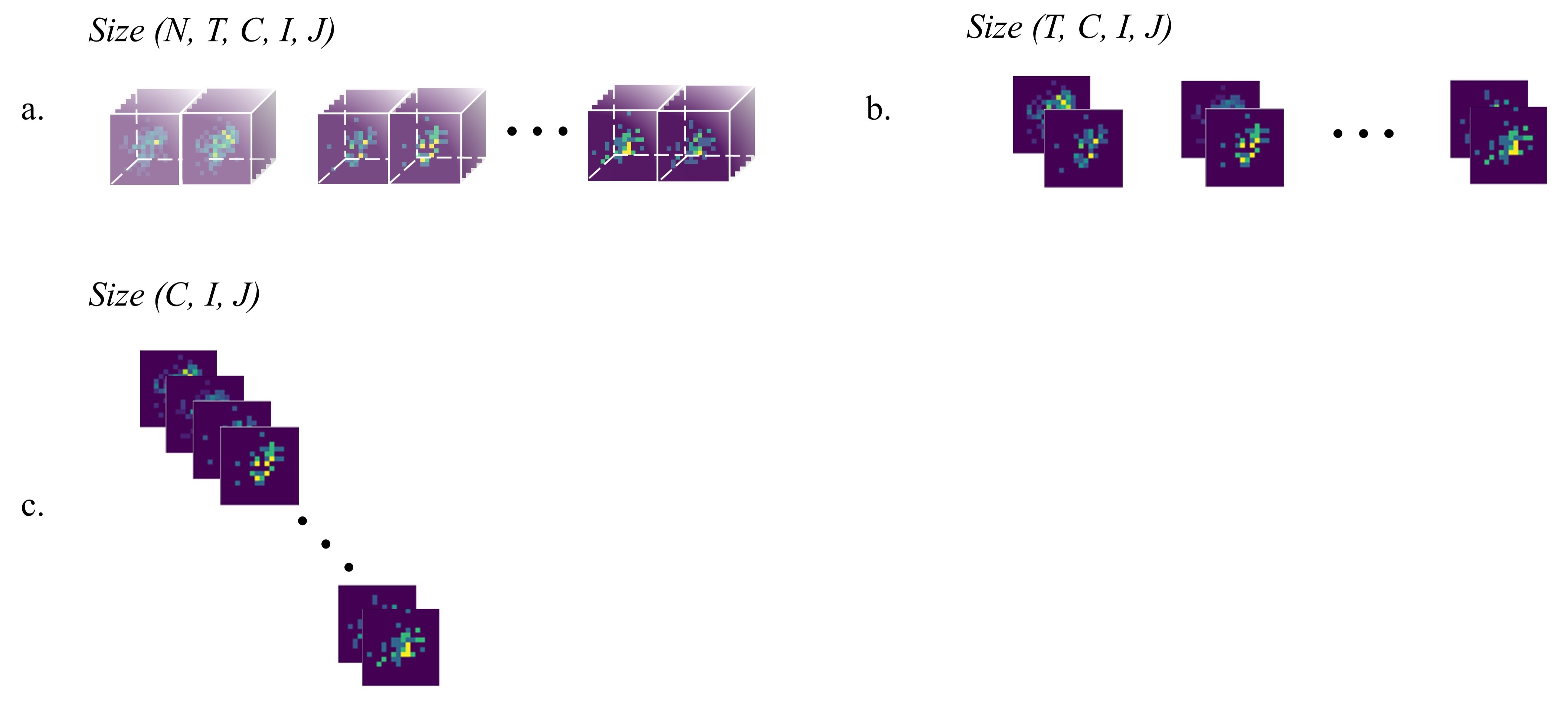}
	\caption{Inputs for different models. $N$ is the fragment, $T$ is the frame in the fragment, $C$ is channel, $I$ is width, and $J$ is height.}
	\label{inputs and outputs}
	\end{figure*}

Our experiments are conducted with an Nvidia V100 16GB GPU.
MSTF-Net consists of 2-layer 3D-CNN encoder blocks, 2-layer E3D-LSTM blocks, 2-layer 3D-CNN decoder blocks, and 2-layer FC blocks.
During the training process of deep learning models, we stop training when the validation error does not improve for consecutive maximum (i.e., 50 epochs in our study) iterations.
%
%
For comparison,
we have selected seven baseline models: MST3D-ResNet, E3D-LSTM, ST-ResNet, 3D-CNN-ResNet, ConvLSTM, 2D-CNN-ResNet, and HA.
For the deep learning models, the number of filters is set to 32, and the size of filter is set to 3.
\begin{enumerate}[(1)]

	\item \textbf{HA}: The traditional time-series model that averages the historical flow to forecast the future flow.
For example, the future flow during 7-8 AM in the grid $(i,j)$ is predicted by averaging the historical flow during 7-8 AM in
	$(i,j)$.
	\item \textbf{2D-CNN-ResNet}: 6-layer residual 2D convolutional neural network.
	Input dimensions of flow videos are shown in \Cref{inputs and outputs} (c).
	\item \textbf{ConvLSTM}: 4-layer convolutional long short-term memory network \citep{xingjian2015convolutional}, which utilizes 2D convolution to extract spatial dependence and LSTM to extract temporal dependence for spatiotemporal data.
	Input dimensions of flow videos are in \Cref{inputs and outputs} (b).
	\item \textbf{3D-CNN-ResNet}: 6-layer residual 3D convolutional neural network \citep{d1}.
	Input dimensions of flow videos are shown in \Cref{inputs and outputs} (b).
	\item \textbf{ST-ResNet}: 12-layer residual 2D convolutional   neural network \citep{zhang2017deep}, which consists of three branches for the closeness, period, and trend properties.
	Input dimensions of flow videos are shown in \Cref{inputs and outputs} (c).
Note that the inputs of 2D-CNN require images without temporal dimension, so the temporal dimensions $N$ and $T$ are squeezed to the channel dimension $C$.
\item \textbf{E3D-LSTM}: 2-layer or 4-layer eidetic 3D convolutional long short-term memory network \citep{d3}.
It consists of same FC blocks with MSTF-Net for processing external factors.
Input dimensions of flow videos are shown in \Cref{inputs and outputs} (a).
\item \textbf{MST3D-ResNet}: 12-layer residual 3D convolutional  neural network \citep{chen2021multiple}, which consists of three branches for the closeness, period, and trend properties. It can be regarded as an improving version of ST-ResNet. 
Input dimensions of flow videos are shown in \Cref{inputs and outputs} (b).

\end{enumerate}




\subsection{Results and Analysis} \label{compare}

For each baseline model, we choose the best performing parameters on the validation set for comparison.
The performance of each model is evaluated
on Root Mean Squared Error (RMSE).
given by
\begin{equation*}
	\mathrm{RMSE}=\sqrt{\frac{1}{n} \sum_{i=1}^{n}(y^{(i)}-\hat{y}^{(i)})^{2}}, 
\end{equation*}
where $y^{(i)}$ is the $i$th real hourly flow, $\hat{y}^{(i)}$ is the $i$th estimated hourly flow, $n$ is the size of test set.


The following experiments results (see details in \Cref{reults}) can be observed:
\begin{enumerate}[(1)]
\item The proposed MSTF-Net performs  best by RMSE. 
\item Increasing layers from 2 to 4 little improves the performance for pure E3D-LSTM models.
\item MSTF-Net significantly performs better than the pure 3D-CNN or E3D-LSTM models such as E3D-LSTM-2, E3D-LSTM-4, MST3D-ResNet, and 3D-CNN-ResNet.



\item 
3D-CNN-ResNet performs the same or better  than 2D-CNN-ResNet, and 
MST3D-ResNet  performs better than ST-ResNet.
In other words, keeping temporal dimension independent, and performing 3D convolution simultaneously on spatial and temporal dimensions significantly improve the performance.
\item 
MST3D-ResNet and ST-ResNet perform worse than 3D-CNN-ResNet and 2D-CNN-ResNet on MoBike. 
We conclude that the design of three branches for the closeness, period, and trend properties is improper for MoBike that only has closeness property.

	\end{enumerate}

\begin{table}[HT]
	\centering
	\caption{The prediction results. The total frames are 36 on MoBike and 18 on NYCBike. E3D-LSTM-2 and E3D-LSTM-4 contain 2 layers and 4 layers, respectively.}
	\label{reults}
	\begin{tabular}{|c|c|c|}
		\hline
		Datasets              & MoBike        & NYCBike       \\ \hline Metrix 
											  & RMSE          & RMSE          \\ \hline
		MSTF-Net & \textbf{1.16} & \textbf{5.59} \\ \hline
		MST3D-ResNet                          & 1.28          & 5.85          \\ \hline
		E3D-LSTM-2                           & 1.21          & 6.03          \\  \hline
		E3D-LSTM-4                          & 1.28          & 6.00          \\  \hline
		ST-ResNet                             & 1.39          & 6.02          \\ \hline
		3D-CNN-ResNet                         & 1.18          & 5.90          \\ \hline
		ConvLSTM                              & 1.18          & 6.53          \\ \hline
		2D-CNN-ResNet                         & 1.22          & 5.90          \\ \hline
		HA                                    & 1.74          & 16.18         \\ \hline
		\end{tabular}
		\end{table}

\subsection{Parameter Sensitivity}\label{parameter sensitivity}
We perform the sensitivity analysis for input length  on NYCBike. In \Cref{se_reults}, we can observe that: 
\begin{table}[ht]
	\centering
	\caption{Parameter sensitivity for input length of frames on NYCBike. }
	\label{se_reults}
	\begin{tabular}{|c|c|c|c|}
	\hline
	Total frames & 18            & 36            & 72            \\ \hline
	Metrix           & RMSE          & RMSE          & RMSE          \\ \hline
	MSTF-Net   & \textbf{5.59} & \textbf{5.56} & \textbf{5.79} \\ \hline
	MST3D-ResNet     & 5.85          & 5.94          & 5.89          \\ \hline
	E3D-LSTM-2      & 6.03          & 6.14          & 6.20          \\ \hline
	ST-ResNet        & 6.02          & 6.00          & 6.25          \\ \hline
	3D-CNN-ResNet    & 5.90          & 5.72          & 5.95          \\ \hline
	ConvLSTM         & 6.53          & 6.10          & 6.20          \\ \hline
	2D-CNN-ResNet    & 5.90          & 5.89          & 6.43          \\ \hline
	\end{tabular}
	\end{table}

\begin{enumerate}[(1)]
\item MSTF-Net still performs best when the input length increases from 18 to 36 and 72. 
\item When the input length dramatically increases from 18 to 72, RMSE rapidly increases for 2D CNN based models: 2D-CNN-ResNet and ST-ResNet. 
For 3D-CNN based models: 3D-CNN-ResNet and MST3D-ResNet, RMSE slightly increases possibly because 3D convolution is more suitable than 2D convolution for capturing long-term spatiotemporal dependence.
\item When the input length increase, almost all models' RMSE increases.
Although longer input contains more information, it also introduces more noises and dramatically increases training time.
ConvLSTM' RMSE decreases because it behaves poorly on short sequences (i.e., 18), and has more potential to reduce on long sequences.
\end{enumerate}

\subsection{Ablation study}\label{ablation study}
\begin{table}[ht]
	\centering
	\caption{Ablation study of external factors. MST-Net excludes the external factors and FC blocks of MSTF-Net. }
	\label{ablation}
	\begin{tabular}{|c|c|c|c|c|}
	\hline
	Datasets      & MoBike        & \multicolumn{3}{c|}{NYCBike}                  \\ \hline
	Num of frames & 36            & 18            & 36            & 72            \\ \hline
	Metrix        & RMSE & RMSE & RMSE & RMSE          \\ \hline
	MSTF-Net      & \textbf{1.16} & 5.59          & \textbf{5.56} & 5.79          \\ \hline
	MST-Net       & 1.18          & \textbf{5.55} & \textbf{5.56} & \textbf{5.57} \\ \hline
	\end{tabular}
	\end{table}
We explore the influence of multiple blocks design in MSTF-Net and observe that: 
\begin{enumerate}[(1)]
\item In \Cref{reults} and \Cref{se_reults}, MSTF-Net that consists of multiple blocks significantly performs better than pure 3D-CNN and E3D-LSTM models on MoBike and NYCBike.
We conclude that 3D-CNN blocks highlight extracting short-term spatiotemporal dependence in each fragment.
The preprocessing of 3D-CNN blocks helps following E3D-LSTM blocks to further extract long-term spatiotemporal dependence over all fragments.

\item In \Cref{ablation}, we show that the fusion of flow videos and external factors has different influences.
Specifically, on a total of four datasets, MSTF-Net performs better on one dataset, MST-Net performs better on two datasets, and they perform the same on one dataset.
\end{enumerate}

\subsection{Visualization}\label{visua}
\begin{figure*}[ht]
	\centering
	\includegraphics[width=0.9\textwidth]{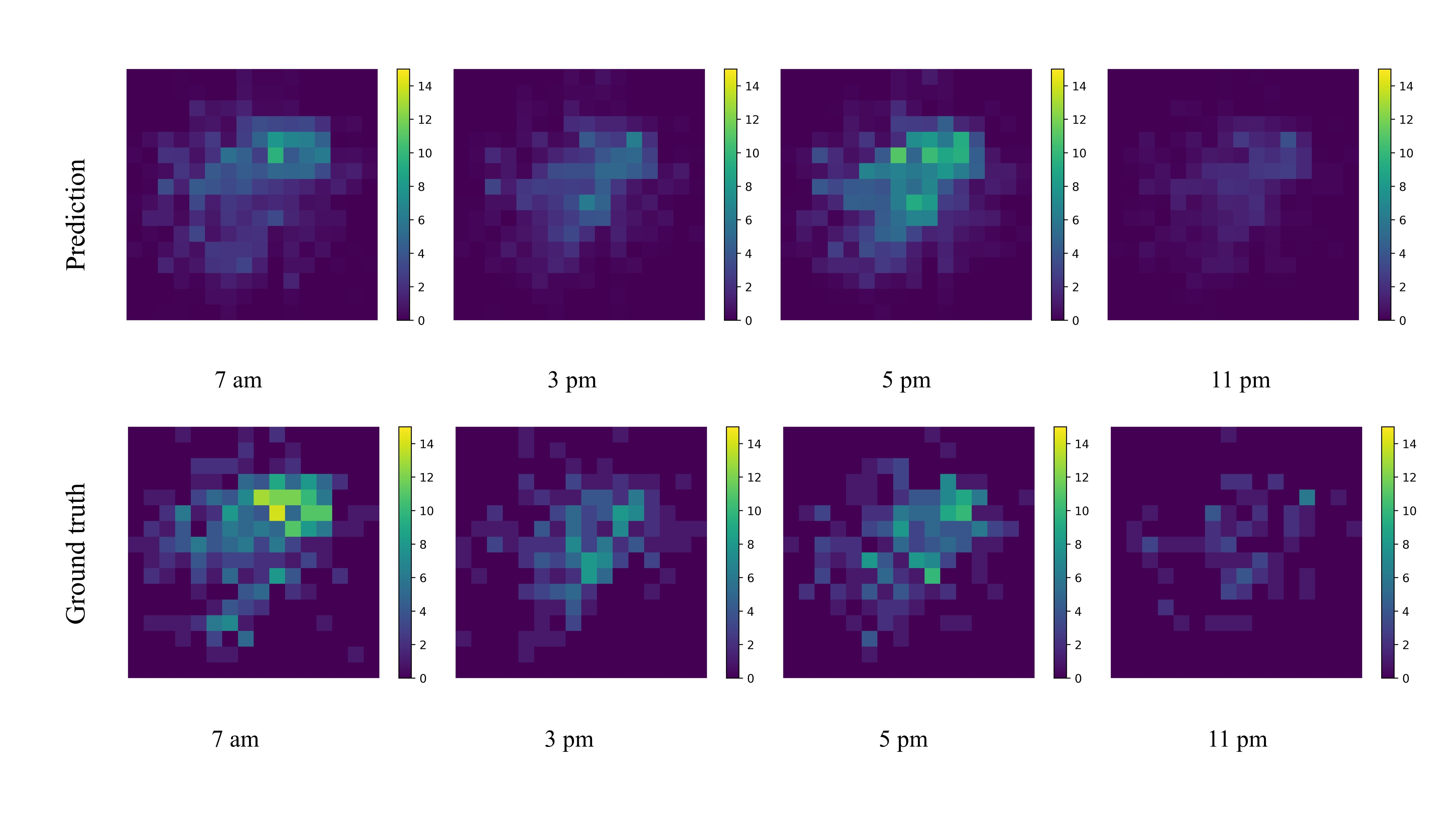}
	\caption{Comparison of the ground truth and prediction by MSTF-Net on MoBike.}
	\label{fig result}
\end{figure*}
Take MoBike as an example, we present some samples of heat maps of the ground truth outflow and predicted results by MSTF-Net, as shown in \Cref{fig result}, where the
brighter color means a larger outflow.
It is obvious that the outflows in morning rush hours and evening rush hours (e.g., 7 am and 5 pm) are
much higher than that in day time and night time (e.g., 3 pm and 11 pm).
The outflow is unbalanced across space:
the central grids are much
higher than the marginal grids.
The trend of the outflow over time is even different in different grids, which makes it hard to predict.
From the samples of visualization, we can find that MSTF-Net 
primarily captures the spatiotemporal characteristics of the outflow.
The combination of
demand forecasting and visualization helps operators to rebalance bikes efficiently.



\section{Conclusions} \label{conclusion}
In this study, a deep learning architecture called MSTF-Net is proposed to forecast bike demands in  bike-sharing systems.
In MSTF-Net, 3D-CNN blocks highlight extracting short-term spatiotemporal dependence in each fragment (i.e., closeness, period, and trend); E3D-LSTM blocks extract long-term spatiotemporal dependence over all fragments; FC blocks extract nonlinear correlations of external factors.
We show that MSTF-Net significantly performs better than the pure 3D-CNN or E3D-LSTM models.

\section*{Acknowledgement}
Gang Kou’s research is has been partially supported by grants
from
the National Natural Science Foundation of China
(U1811462, 71725001, and 71910107002),
State Key R \& D Program of China (2020YFC0832702), and
Major Project of the National Social Science Foundation of China (19ZDA092).
Feng Xiao's research has been partially supported by grants from
the National Science Fund for Distinguished Young Scholars (72025104) and
the National Natural Science Foundation of China (71861167001).
Xianghua Gan's research is supported by
National Natural Science Foundation of China (71771189).

\bibliographystyle{cas-model2-names}

\bibliography{fed}

\appendix
\section{Appendix}\label{app}
	The spatiotemporal part of LSTM unit at time stamp $t$ and layer $k$ are shown as follows:

		\begin{align}
		&i_{t}=\sigma\left(W_{x i} * \mathcal{X}_{t}+W_{h i} * \mathcal{H}_{t-1}^{k}+b_{i}\right) \\
		&g_{t}=\tanh \left(W_{x g} * \mathcal{X}_{t}+W_{h g} * \mathcal{H}_{t-1}^{k}+b_{g}\right) \\
		&f_{t}=\sigma\left(W_{x f} * \mathcal{X}_{t}+W_{h f} * \mathcal{H}_{t-1}^{k}+b_{f}\right) \\
		&i_{t}^{\prime}=\sigma\left(W_{x i}^{\prime} * \mathcal{X}_{t}+W_{m i} * \mathcal{M}_{t}^{k-1}+b_{i}^{\prime}\right) \\
		&g_{t}^{\prime}=\tanh \left(W_{x g}^{\prime} * \mathcal{X}_{t}+W_{m g} * \mathcal{M}_{t}^{k-1}+b_{g}^{\prime}\right) \\
		&f_{t}^{\prime}=\sigma\left(W_{x f}^{\prime} * \mathcal{X}_{t}+W_{m f} * \mathcal{M}_{t}^{k-1}+b_{f}^{\prime}\right) \\
		&\mathcal{C}_{t}^{k}=i_{t} \odot g_{t}+f_{t} \odot \mathcal{C}_{t-1}^{k} \\
		&\mathcal{M}_{t}^{k}=i_{t}^{\prime} \odot g_{t}^{\prime}+f_{t}^{\prime} \odot \mathcal{M}_{t}^{k-1} \\
		&o_{t}=\sigma(W_{x o} * \mathcal{X}_{t}+W_{h o} * \mathcal{H}_{t-1}^{k}+W_{c o} * \mathcal{C}_{t}^{k}+ \\ &W_{m o} * \mathcal{M}_{t}^{k}+b_{o}) \\
		&\mathcal{H}_{t}^{k}=o_{t} \odot \tanh \left(W_{1 \times 1} *\left[\mathcal{C}_{t}^{k}, \mathcal{M}_{t}^{k}\right]\right),
		\end{align}

		where $\sigma$ is the sigmoid function, $\ast $ is the convolution operator, and $\odot $ denotes the Hadamard product.
		There are four inputs:
    $\mathcal{X}_{t}$, the raw frame or hidden states from the previous layer;
$\mathcal{M}_{t}^{k}$, the
		previous spatiotemporal memory;
$\mathcal{H}_{t-1}^{k}$
		and $\mathcal{C}_{t-1}^{k}$, the previous hidden states and memory states.
		Two sets of gate structures, including input gate $i_t$ and $i_t ^\prime$, forget gate $f_t$ and $f_t ^\prime$, as well as the
		output gate $o_t$, control the information flow in space-time domain.
All of them can be presented by
		$\mathcal{R}^ {H \times W \times C }$ dimensional tensors, where the first two dimensions are the width and height of feature
		maps, and the last one is the number of feature map channels.

\end{document}